\documentclass[numbib]{imaiai}
\usepackage[utf8]{inputenc}
\usepackage{graphicx}
\usepackage{color}

\begin{document}

\title{Multi-hop assortativities for network classification}

\shorttitle{Multi-hop assortativities for networks classification} 
\shortauthorlist{L. GUTIERREZ, J-C. DELVENNE} 

\author{{
\sc Leonardo Gutiérrez-Gómez}$^*$,\\[2pt]
Institute for Information and Communication Technologies, Electronics and Applied
Mathematics (ICTEAM), Université catholique de Louvain, Avenue Georges Lemaître, 4,
1348 Louvain-la-Neuve, Belgium\\
$^*${\email{Corresponding author: leonardo.gutierrez@uclouvain.be}}\\[2pt]
{\sc and}\\[6pt]
{\sc Jean-Charles Delvenne} \\[2pt]
Institute for Information and Communication Technologies, Electronics and Applied
Mathematics (ICTEAM) and Center for Operations Research and Econometrics (CORE), Université catholique de Louvain, Avenue Georges Lemaître, 4,
1348 Louvain-la-Neuve, Belgium \\
{jean-charles.delvenne@uclouvain.be}}

\maketitle

\begin{abstract}
{Several social, medical, engineering and biological challenges rely on discovering the functionality of networks from their structure and node metadata, when it is available. For example, in chemoinformatics one might want to detect whether a molecule is toxic based on structure and atomic types, or discover the research field of a scientific collaboration network.

Existing techniques rely on counting or measuring structural patterns that are known to show large variations from network to network, such as the number of triangles, or the assortativity of node metadata. We introduce the concept of multi-hop assortativity, that captures the similarity of the nodes situated at the extremities of a randomly selected path of a given length. We show that multi-hop assortativity unifies various existing concepts and offers a versatile family of  `fingerprints' to characterize networks. These fingerprints allow in turn to recover the functionalities of a network, with the help of the machine learning toolbox.

Our method is evaluated empirically on established social and chemoinformatic network benchmarks. Results reveal that our assortativity based features are competitive providing highly accurate results often outperforming state of the art methods for the network classification task.}
\\
{network classification, multi-hop assortativities, graph classification}
\end{abstract}

\section{Introduction}
One of the early tasks of network science has been to characterize complex networks through a few global characteristics that summarize their structure in order to understand how different networks, from different fields or with different functional properties behave \cite{Newman03thestructure}. For instance it is well known that many social networks are highly assortative, as people tend to make social connections with other that are similar e.g in terms of age, race or degree in the network. On the other hand, various biological networks tend to be dissassortative as for instance food web networks \cite{Dunne12917}. The clustering coefficient, measuring the
density of triangles, is also shown to be distinctive characteristics of many social networks.
\begin{figure}[!ht]
\centering{\includegraphics[width=4.5in]{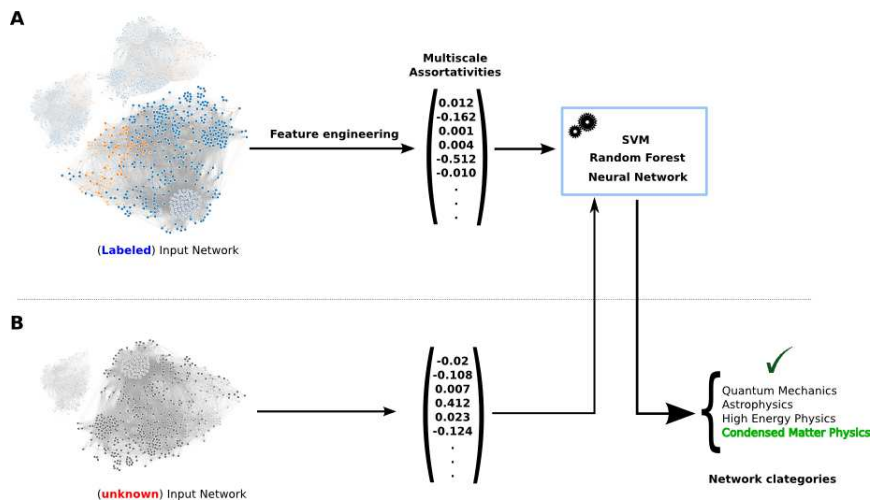}}
\caption{\textbf{Overview of the proposed approach}. Given a set of networks, we aim to predict the most likely  class to which each network belongs to. Feature vectors of multi-hop assortativities coefficients are built through a feature engineering process. (A) In the training phase, a subset of labeled networks (for which we know the true class) are used to learn a classifier. They can have nodes metadata (represented by node colors on the picture) or not. (B) The class to which an unseen network belongs is predicted by feeding the trained model with the feature vector representation. } 
\label{diagram}
\end{figure}

As the amount of available data on networks keeps growing, one is tempted to use such features systematically in order to correlate them with the context or function attached to the network, using the tools of machine learning, statistics or data mining. One typical task in this context is supervised classification, where we seek to predict for example the toxicity or anti-cancer activity of molecules based on its structure, or which field of physics a researcher contributes to, based on their collaboration ego-network. We speak of supervised classification because an optimal classification criterion is built automatically from a \textit{training} dataset for which the correct answer is known. 

In order to apply the powerful supervised classification techniques that have been developed in the last decades (support vector machines, random forests, artificial neural networks, etc.), we might to perform a crucial feature engineering process which can be achieved with one of these two strategies. The first one is to transform a given network into a fixed-dimensional vector of numbers characterizing the network (those numbers are called the features of the network); every network now becomes a vector in a Euclidean space (the feature space) so that any classification technique can be used in order to learn which regions of the feature space correspond to toxic molecules, or to High Energy Physicists collaboration networks, etc. We call it the feature-based approach.
The second strategy defines a quantitative measure of similarity between pair of networks, called a kernel. This kernel can often be seen as the scalar product between implicit high-dimensional feature representations of a network. This so-called kernel trick allows to classify networks without ever computing explicitly the coordinates of data points in the high-dimensional feature space, sometimes with a substantial gain of computational time over a high-dimensional feature-based approach.

In this paper we build on the idea that a dynamical process on a network is strongly indicative of
the structure of the underlying graph in which the process takes place, an idea used e.g. in community
detection \cite{Delvenne2013a} or node centrality \cite{Brin1998}. Here, the main idea is to use features associated to the random walk
dynamics as network descriptors. Those features are obtained in a systematic way from the correlation patterns of node attributes seen by a random walker at different time instants. The attributes can be
either structural (for instance eigenvectors, including the degree) or metadata of the node (e.g. atom type
in a molecule network, age or gender in a social network). While such correlation between two
consecutive nodes visited by the random walker coincides with the usual assortativity coefficient, at
different times scales one obtains a sequence of \textit{multi-hop assortativities} describing the patterns of
average similarity between a node and larger and larger neighborhoods. 

The eigenvectors, invaluable source
of information when it comes to characterizing a dynamical system, have long be used to explore the
structure of the network in terms of centrality (PageRank, degree, eigenvector centrality) or community
detection (spectral methods), and thus form a natural suspect for investigating correlation patterns. Moreover,
considering now the ID of the node as its attribute (each node being a category on its own), one will see
that a high three-hop assortativity (essentially the probability that a three-hop random walk from a node comes back to the node) is related to the presence of many triangles in the network, hence is
akin to a clustering coefficient. The decay to zero of the multi-hop assortativities as the number of hops
increases is indicative of the mixing time of the random walker, also indicative of the ‘small-worldness’
of the network and its heterogeneous structure (existence of communities).

We claim therefore that the multi-hop assortativities generated by a random walker offer a natural,
conceptually simple way to generate systematic features acting as a numerical fingerprint for a network,
from which an efficient feature-based classification of networks can be performed, see Fig. \ref{diagram}. It is flexible in that it embeds well known
notions (density of triangles, degree assortativity, etc) and allows to use categorical (gender, atom type,
etc.) or scalar (age, weight, etc) metadata if available within the same framework.

We evaluate our approach experimentally on social networks and chemoinformatics benchmarks datasets. We compare its classification accuracy with respect to some representative graph kernels, neural networks and features based algorithms of the literature. To make statistical inference from the observed difference in accuracies, we performed a series of simple and robust non-parametric tests for comparison of the algorithms \cite{Demsar2006}. The accuracy measures are therefore  compared using the Friedman's average rank test, and Nemenyi's post hoc test will be employed to test the significance of the differences in rank between individual algorithms.

Our results reveal that generalized
assortativities on a small set of elementary structural network features, and any exogenous node metadata (if available), are capable of achieving and outperform in many cases state-of-the-art accuracies, with reasonable computational resources.

The paper is structured as follows: Section 2 reviews related approaches in the literature. Section
3 introduces multi-hop assortativities. Section 4 addresses the problem of definition of features of
network dynamics. Finally, in sections 5 and 6 we present our experimental setting in which we assessed
our method and discussion.

\section{Related work}
Graph classification has been extensively studied by the network science and machine learning communities under different perspectives. Methods can be grouped in feature based models and graph kernels methods approaches. We briefly introduce the methods against which we will compare our methodology. 

There is a considerable amount of literature related with graph kernel-based methods. They can be categorized in three classes: graph kernels based on \textit{random walks} \cite{Borgwardt}, \cite{DBLP:conf/icml/KashimaTI03}, kernels based on \textit{subtree patterns}, \cite{Ramon03expressivityversus}, \cite{Shervashidze2009FastSK}, \cite{Shervashidze:2011:WGK:1953048.2078187},  and also kernels based on \textit{limited-size subgraphs} or \textit{graphlets}, \cite{5664}, \cite{Horvath2005}. The similarity between graphs is assessed by counting the number of common patterns, or decomposing the input graphs into substructures such as shortest path or neighborhood subparts. However, kernels based on such substructures are computationally expensive, sometimes even NP-hard to determine and also limited in expressiveness.  Moreover, the complexity of kernel computation for all pairs of networks in the training phase grows quadratically in the number of examples.

Improved graph kernels such as Weisfeiler-Lehman (WL) Subtree Kernel \cite{Shervashidze2009FastSK}, and Neighborhood Subgraph Pairwise Distance Kernel \cite{267297} scale better by defining similarity between a restricted, easy-to-compute class substructures. A recent work \cite{Yanardag:2015:DGK:2783258.2783417} proposes a deep version of the Graphlet, WL and Shortest-Path kernels. Patterns in the network are transformed into features in the same way that words are embedded in a Euclidean space in a natural language processing model, with so-called CBOW or Skip-gram algorithms \cite{journals/corr/abs-1301-3781}.

Automatic feature learning algorithms aim to learn the underlying graph patterns often through a neural network variant. More recently \cite{Niepert2016} proposes to learn a convolution neural network (PSCN) for arbitrary graphs, by exploiting locally connected regions directly from the input graph. Shift aggregate extract network  (SAEN) \cite{DBLP:journals/corr/OrsiniBF17} introduces a neural network to learn graph representations, by decomposing the input graph hierarchically in compressed objects, exploiting symmetries and reducing memory usage.

The spectrum of graphs as feature for graph classification has been explored by \cite{Schmidt2014}, \cite{Wilson:2005:PVA:1070615.1070795} who compute features from the spectral decomposition of the Laplacian matrix. Barnett \cite{Barnett2016} proposes a hybrid feature-based approach. It combines manual selection of network features with existing automated classification methods. 

Our method can be seen as a versatile feature-based model that create multiscale patterns from any node attribute, whether they are structural, spectral or exogenous  (metadata), whether they are numerical or categorical. This is in contrast for example  with  Graphlet \cite{5664}, Deep Graphlet (DGK)\cite{Yanardag:2015:DGK:2783258.2783417} or Feature-Based (FB) \cite{Barnett2016} methods who cannot exploit node metadata. The feature vector representation of a network is created using multi-hop assortativity patterns from a random walker perspective. In this work we show that even a very limited set of structural or spectral node attributes can be `amplified' to a rich network feature representation allowing performant classification.


\section{Random walks and multi-hop assortativities}

In this section we introduce multi-hop assortativities through a random walk dynamic.

Consider an undirected, connected and non-bipartite graph $G=(V,E)$ with $N$ vertices and $m$ edges. We will assume for simplicity  that the graph is unweighted, but all the results are applicable to the nonnegative weights. Extension to directed networks is straightforward as well. The adjacency matrix $A$ associated to $G$ is an $N \times N$ binary matrix, with $a_{ij}=1$ if the vertices $i$ and $j$ are connected, and $0$ otherwise. The number of edges in the vertex $i$ is known as the \textit{degree} of the vertex $i$, denoted $d_i$. Then the degree vector of $G$ can be compiled as $\textbf{d} = A \textbf{1}$, where \textbf{1} is a $N \times 1$ ones vector. For posterior computations, we define also the $N \times N$ degree matrix  $D=diag(d)$.

The standard random walk on such a graph defines a Markov chain in which transition probabilities are split uniformly among the edges, with a transition probability $1/d_i$:

\begin{equation}\label{markov}
\textbf{p}_{t+1} = \textbf{p}_t [D^{-1}A ] \equiv \textbf{p}_t M
\end{equation}

Here $\textbf{p}_t$ is the $1 \times N$ normalized probability distribution of the random walker over the nodes at time $t$, and $M$ the $N \times N$ transition matrix. Under the assumptions on the graph (connected, undirected, and non-bipartite), the dynamics converges to a unique \textit{stationary distribution} vector $\pi = d^T/2m$.
We define also the $N \times N$ diagonal matrix $\Pi = diag(\pi)$. 

Consider a scalar node attribute such as the degree, centrality, age (in a social network), etc. It is known that many real networks exhibit interesting assortativity properties, i.e. a high covariance between the two end node's attributes of a randomly selected edge. For the random walker in stationary distribution, this translates into the covariance of the node attribute visited at two consecutive time instants $\tau$ and $\tau+1$, since each edge of the graph is visited by the walker with the same frequency. 

\begin{figure}[!t]
\centering{\includegraphics[width=5.5in]{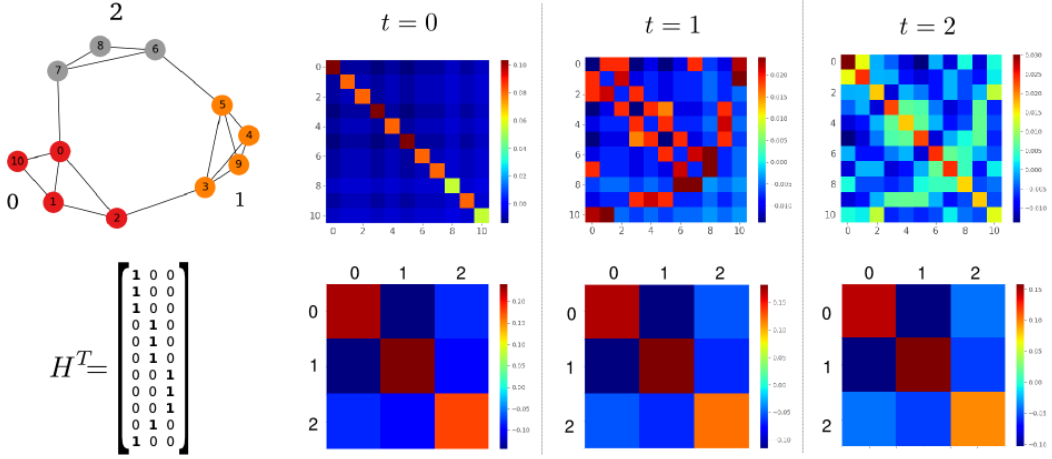}}
\caption{\textbf{Autocovariance of a categorical attribute.} Top matrices correspond to the autocovariance matrices (Eq. \ref{covariance}) of a random walk dynamics on the left hand graph. Categorical node attributes are represented by colors on the graph. They define a partition encoded in the binary matrix $H$. Bottom matrices correspond to the category-level autocovariance matrix $H^T \rho(t)H$ for different time instants. In particular, the diagonal entries are $r(t, h_i)$, see Eq. \ref{diagfeature}, and the trace of the matrix is $r(t,H)$, see Eq. \ref{autocovarience}.} 
\label{fig_covariance}
\end{figure}

This suggests to consider \textit{multi-hop assortativities}, as the covariance between nodes attributes seen by the walker at times $\tau$ and $\tau + t$, for any $t=0,1,2,3,\ldots$. While the case $t=0$ is simply the variance, the case $t>1$ explores assortativity patterns within multi-hop neighborhoods, i.e., the covariances between extremities of a randomly selected path of length $t$.

More formally, we consider the $N-$dimensional vector $X(t)$ as a random indicator vector of the presence of a random walker in time such that for the $k$th entry $X_k(t) = 1$ if the walker is at node $k$ at time $t$, and zero otherwise. Thus, the \textit{autocovariance} matrix \cite{Delvenne2013a} of the observable process at time $t$  is expressed as
\begin{equation}\label{covariance}
\rho (t) = \text{Cov}(X_{\tau}, X_{\tau + t}) = \Pi M^t - \pi^T \pi.
\end{equation}
Compiling the scalar nodes attributes in a $N \times 1$ attribute vector $v$, the covariance of the attribute $v$ is given by

\begin{equation}\label{vector_covariance}
r(t,v) =  v^T \rho(t) v
\end{equation}
as indeed the value of the node attribute observed by the random walker on the node it stands on at time $\tau$ is none but the random variable $\sum v_k X_k(\tau)$ summed over all nodes $k$. \\

We call $r(t, v)$ the $t$-hop \textit{assortativity} of a scalar attribute $v \in \mathbb{R}^{N \times 1}$. As 
mentioned above, for $t = 1$ on an undirected network this coincides with the usual assortativity of a scalar attribute, except that the latter is often normalized as a correlation. We choose not to normalize it as it increases the dynamical range of the assortativity, hence its discriminating power to distinguish a network from another. For $t = 0$ it is simply the variance of attribute $v$ among the nodes, with each node's attribute weighted proportionally to the degree of the node. 

Note that as the number $t$ of hops grows to infinity, $\rho(t)$ converges to the zero matrix for a connected non-bipartite network, as every row of $M^t$ converges to $\pi$. Therefore all multi-hop assortativities tend to zero with the number of hops, and
the rate of convergence is well-known to be given by the spectral gap of $M$. The spectral gap is the inverse of the mixing time of the random walk, indicative of an effective typical diameter of the graph, and equivalent through Cheeger's inequality the existence of a strong community structure in the network \cite{ALON198573,chung2005laplacians}. Note that if the network is connected but bipartite, a unique stationary vector $\pi$ is still defined and $\rho(t)$ can still be constructed, but oscillates periodically as the number $t$ of hops grows.

It often happens that a node attribute is categorical rather than numerical, e.g. gender or political parties in a social network or atom type in chemical compounds. In this case we compute an assortativity coefficient by encoding each category as a binary characteristic vector. One therefore encode  
a $k$-category attribute as an $N \times k$ binary matrix $H$, every row which contains exactly one  $1$ (one-hot encoding). We compute the autocovariance of each category, i.e. each column $h_i$ of $H$ as

\begin{equation}\label{diagfeature}
r(t,h_i) = h_i^T\rho(t) h_i 
\end{equation}
with $1 \leq i \leq k $, yielding a $k$-dimensional vector for each time $t$.

It can also be written directly in terms of probability as follows \cite{Lambiotte2014a, Delvenne2013a} 

\begin{equation} 
r(t,h_i) = \textrm{Proba}(\textrm{$x_{\tau}$ and $x_{\tau+t}$ both in category $i$}) - \textrm{Proba}(\textrm{$x_{\tau}$ and $x_{\tau+\infty}$ in category $i$})
\end{equation} 
where $x_\tau$ is the node visited by the random walker at time $\tau$. In other words, it measures how likely it is that a randomly selected node and a randomly selected $t$-hop neighbour both belong to category $i$. We remove the term corresponding to infinite delay $t \to \infty$, which is also the probability $(\pi h_i)^2$ that two independent random walkers belong to category $i$.



In the case of many categories, it is convenient to
characterize the autocovariance of a given node labeling with a given time delay with a single number,
and therefore we sum the autocovariances of each category:

\begin{equation} \label{autocovarience}
r(t,H) = \sum_{i=1}^k r(t,h_i) = \text{Tr}[H^T \rho(t)H)]
\end{equation} 

It can also be written directly in terms of probability as

\begin{equation} 
r(t,H) = \textrm{Proba}(\textrm{$x_t$ and $x_{\tau+t}$ in same category}) - \textrm{Proba}(\textrm{$x_t$ and $x_{\tau+\infty}$ in same category})
\end{equation} 
In other words, it measures how frequently a randomly selected node and a randomly selected $t$-hop neighbor belong to the same category.  We remove the term corresponding to infinite delay $t \to \infty$, which is also the probability $\|\pi H\|_2^2$ that two independent random walkers belong to the same category, which is $\sum_{i} (\pi h_i)^2$.

For $t = 1$ this is the usual, unnormalized assortativity coefficient of the categorical attribute \cite{doi:10.1177/001316446002000104}. For $t = 0$, this coefficient does not depend on the structure of the network, only
of the relative frequencies of the categories, taking a high value $1 - \dfrac{1}{k}$ if the $k$ categories are equally
spread, in terms of total degrees of nodes of each category. Uneven categories lead to a variance close
to zero. This measure of diversity is essentially the Rényi entropy of the node attribute \cite{Gray:1990:EIT:90455}, or Simpson's
diversity index used in ecology or economics \cite{simpson}. For general $t$, this coefficient is the \textit{multi-hop assortativity} of the categorical node attribute encoded by the matrix $H$, and captures the tendency of nearby nodes to having the same categorical attribute.

It can be noted that a categorical attribute $H$ defines a partition of the nodes, and $r(1,H)$ happens to coincide with the modularity of the partition \cite{PhysRevE.74.036104} used in community detection, or graph clustering. The quantity $r(t,H)$, called in this context Markov stability \cite{Delvenne2013a}, allows multi-scale community detection by scanning through the time parameter $t$.  
\begin{figure}[!ht]
\centering{\includegraphics[width=4.6in]{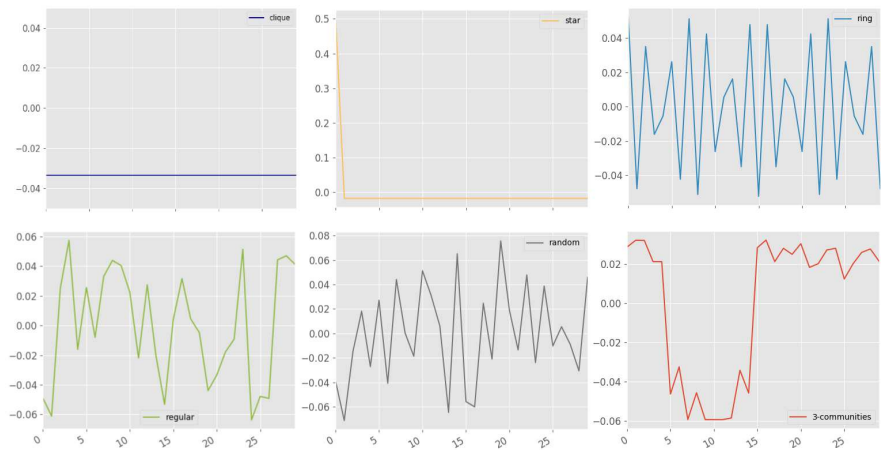}}
\caption{\textbf{Eigenvectors as network features.} We plot the second dominant eigenvector ($\pi_2$) of M of six different network topologies with 30 nodes each: (1) clique,  (2) star, (3) ring, (4) random $8$-regular graph (a graph picked randomly uniformly among all undirected graphs with all nodes of degree 8), (5) Erd\H{o}s-R\'enyi random graph with $p=0.4$ (6) network with a planted partition of 3 communities with 5, 10 and 15 nodes, within-group probability of $0.8$ and between-group probability of $0.02$.} 
\label{secondeigenvector}
\end{figure}
\begin{figure}[!ht]
\centering{\includegraphics[width=4.6in]{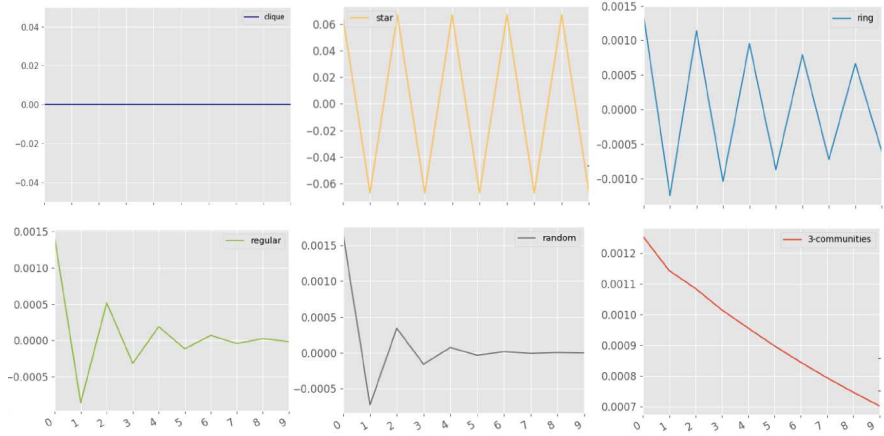}}
\caption{\textbf{Autocovariances as network features.} We plot $r(t,v)$ (Eq. \ref{vector_covariance}) with $v$ as the second dominant eigenvector ($\pi_2$) of the transition matrix M vs time, on six different network topologies with 30 nodes each: (1) clique,  (2) star, (3) ring, (4) random $8$-regular graph (a graph picked randomly uniformly among all undirected graphs with all nodes of degree 8), (5) Erd\H{o}s-R\'enyi random graph with $p=0.4$ (6) network with a planted partition of 3 communities with 5, 10 and 15 nodes, within-group probability of $0.8$ and between-group probability of $0.02$.} 
\label{stab_secondeigenvector}
\end{figure}

\section{Features of network dynamics} \label{features_network}
We have seen how to define multi-hop assortativity coefficients through a random walk dynamic on networks. We argue that multi-hop assortativity patterns of structural and metadata (if available) node attributes emerge as useful fingerprint of the network, characterizing the interaction of nodes attributes
in many levels and discriminating between diverse network structures. We aim to build an expressive feature vector representation of networks based on covariances of nodes attributes. 

Given that we have on top of a network a stationary random walk dynamics, it is natural to consider its spectrum. Indeed, the eigenvectors of  matrices describing random walks on graphs have been used extensively for many graph mining 
purposes, i.e spectral clustering \cite{Kannan:2004:CGB:990308.990313, Ng:2001:SCA:2980539.2980649}, eigenvector centrality \cite{citeulike:1143909}, Pagerank centrality \cite{Brin1998} and so on.  
Moreover, the PageRank, the first left eigenvector of the random walk matrix $M$, coincides in this undirected case with the degree (normalized by the sum of all degrees), and the degree assortativity has been one of the earliest and most important structural properties considered in network science \cite{Newman03thestructure}. It is therefore natural to check if multi-hop assortativities of several leading eigenvectors constitute a good set of features characterizing the structure of the graph. 

Let us illustrate this fact on one example. As can be seen on Figure \ref{secondeigenvector}, choosing a network attribute such as  the second left eigenvector of $M$, yields a reasonable signature to discriminate among different network structures. However, its assortativity, see Figure \ref{stab_secondeigenvector}, has additional advantages: it is an adequate fingerprinting of the network in terms of structural differentiation among topologies but also is a lower dimensional representation of the same network. In Figure \ref{stab_secondeigenvector} each dimension correspond to an assortativity coefficient of the selected node attribute (second dominant eigenvector of $M$) in a given scale (Eq. \ref{vector_covariance}). 

The simplest node attribute we can think about is certainly the node ID  itself, i.e. $H=I$ (identity matrix) in Eq (\ref{autocovarience}). It is straightforward that for $t=0$  it yields an assortativity $r(0,I)=1- \|\pi\|^2$, delivering effectively a variance of the degree distribution. For $t=3$, it counts essentially the probability for the random walker to follow a triangle, thus representing a variant of the global clustering coefficient, yet another structural feature of interest. For arbitrary $t > 0$ it counts essentially the density
of $t$-cycles.

In addition to second-moment measures given by multi-hop assortativities, it is of course natural to include first moments, i.e. averages, in the list of features able to characterize a network. Any scalar
attribute $v \in \mathbb{R}^{N \times 1}$ generates the network feature $\pi^T v = \sum_k \pi_k v_k$ (average weighted by the degrees), while a categorical attribute encoded by the membership matrix $H$ generates the list of respective frequencies of each category, $\pi^T H \in \mathbb{R}^{1 \times k}$.

Certainly real life networks are more complex and heterogeneous that our toy example. In complex networks, is natural to find diversity in mixing patterns arise from diverse nodes attributes. For our experiments, we opt for selecting a reduced set of features capturing local and global structural mixing properties as well as metadata node attributes:

\begin{enumerate}
\item multi-hop assortativities of node IDs, setting $H=I$ in Eq (\ref{autocovarience}) for $0,1, \ldots, t$ hops
\item Average of first $p$ dominant left eigenvector of $M$, i.e $\{\pi^T v: v \text{  is a dominant eigenvector of }M \}$ 
\item multi-hop assortativities of the first $p$ dominant left eigenvectors of $M$ (Eq. \ref{vector_covariance}) for $0,1, \ldots, t$ hops
\item (If available) average of categorical metadata node attributes: $\{ \pi^T h_i : H=[h_1,h_2,..h_k], 1 \leq i \leq k\}$
\item (If available) multi-hop assortativities of categorical metadata node attributes (Eq \ref{diagfeature}) for $0,1, \ldots, t$ hops
\item Number of nodes
\item Number of edges
\end{enumerate}
We compile the aforementioned features in a single feature vector that will be used as a network fingerprint for data mining purposes. Its dimension varies according with the choice of eigenvectors $p$, number of different node categories $k$ the dataset has and the number $t$ of hops for multi-hop assortativities. Our experiments were performed with $t=0,1,2,3$, keeping the first three dominant left eigenvectors of $M$.
For clarity, we emphasize that features number 1, do not require the networks under consideration to share the same size or a particular node ordering, as assortativity of a given number of hops is a single number summing the individual node assortativities, independently of node names or ordering.
 
In the next section we evaluate empirically our method on many real life network datasets and compare against other approaches of the literature.

\section{Experiments and Results}
The aim of our experiments is to show that our multi-hop assortativities are useful descriptors for real life networks discriminating well between classes. Experiments are conducted on networks with and without node metadata. Classification consists in predicting the most likely label to an unseen input example. To do so, we train two popular classifiers widely used in the literature: Support Vector Machines (SVM) \cite{Smola04atutorial} and Random Forest \cite{Breiman2001} (RF). The former consists in learning an optimal hyperplane maximizing the margin of separation between the data points in the feature space.  On the other hand, random forest classifier creates a set of decision trees predictors from randomly selected subset of training set. It aggregates the votes from different decision trees to decide the final class of a test example.

Details about the datasets and experimental setup are explained in the following subsections.
\begin{table}[!h] 
\scriptsize
\caption{Statistics for benchmark datasets. First six correspond to chemoinformatic and the rest to social network datasets. UN means unlabeled nodes \vspace*{0.5cm}}\label{datasets}
\centering
\begin{tabular}{l@{\quad}c@{\quad}c@{\quad}c@{\quad}c@{\quad}c}
\hline
Dataset & Number of graphs & Classes & Node attributes & Average nodes & Average edges \\ 
\hline\rule{0pt}{10pt} 
MUTAG & 188 & 2 & 7 & 17.93 & 19.79 \\ [2pt]
 
PTC & 344 & 2 & 19 & 14.29 & 14.69 \\ [2pt]
 
NCI1 & 4110 & 2 & 22 & 29.87 & 32.3 \\ [2pt]
 
NCI109 & 4127 & 2 & 19 & 29.68 & 32.13 \\ [2pt]

ENZYMES & 600 & 6 & 3 & 32.63 & 62.14 \\ [2pt]

PROTEINS & 1113 & 2 & 3 & 39.06 & 72.82 \\ [2pt]
\hline\rule{0pt}{10pt} 
COLLAB & 5000 & 3 & UN & 74.49 & 2457.78 \\ [2pt]

REDDIT-BINARY & 2000 & 2 & UN & 429.63 & 497.75 \\[2pt] 

REDDIT-MULTI-5K & 5000 & 5 & UN & 508.52 & 594.87 \\ [2pt]

REDDIT-MULTI-12K & 11929 & 11 & UN & 391.41 & 456.89 \\ [2pt]

IMDB-BINARY & 1000 & 2 & UN & 19.77 & 96.53 \\ [2pt]

IMDB-MULTI & 1500 & 3 & UN & 13.0 & 65.94 \\ [2pt]
\hline \rule{0pt}{0pt} 
\end{tabular} 
\end{table}
\subsection{Datasets}
Twelve real-world datasets were used in our experiments. We summarize their general properties in Table  \ref{datasets}. The first six correspond to the chemoinformatic category, composed by undirected graphs representing either molecules, enzymes or proteins. They have all categorical node attributes encoding a particular functional property related its function. The later six are unlabeled node (i.e without node attribute) social network datasets. We will describe them below.

\textbf{MUTAG} \cite{debnath_compadre_debnath_shusterman_hansch_1991} is a nitro compounds dataset divided in two classes according to their mutagenic activity on bacterium \textit{Salmonella Typhimurium}. \textbf{PTC} dataset \cite{helma_king_kramer_srinivasan_2001} contains compounds labeled according to carcinogenicity 
on rodents, divided in two groups. Vertices are labeled by 19 atom types. The \textbf{NC11} and \textbf{NCI109} \cite{wale_karypis_2006} from the National Cancer Institute (NCI), are two balanced dataset of chemical compounds  screened for activity against non-small cell lung cancer and ovarian cancer cell. They have 22 and 19 categorical  node labels respectively. \textbf{PROTEINS} \cite{borgwardt_ong_schonauer_vishwanathan_smola_kriegel_2005}, \cite{schomburg_2004} is a two-class dataset in which nodes are secondary structure elements (SSEs). Nodes are connected if they are contiguous in the aminoacid sequence. \textbf{ENZYMES} \cite{borgwardt_ong_schonauer_vishwanathan_smola_kriegel_2005}, \cite{schomburg_2004} is a dataset of protein tertiary structures consisting of 600 enzymes from the BRENDA enzyme database. The task is to assign each enzyme to one of the 6 EC top-level classes. \\ 

From the social networks pool \cite{Yanardag:2015:DGK:2783258.2783417}, \textbf{COLLAB} is a scientific collaboration dataset, where ego-networks of researchers that have worked together are constructed. The task is to determine whether the ego-collaboration network belongs to any of three classes, namely, \textit{High Energy Physics}, \textit{Condensed Matter Physics} and \textit{Astro Physics}. \textbf{REDDIT-BINARY}, \textbf{REDDIT-MULTI-5K} and \textbf{REDDIT-MULTI-12K} are three balanced datasets having two, five and eleven groups respectively. Each one contains a set of graphs representing an on-line discussion thread where nodes corresponds to users and there is an edge between them if anyone responds to another's comment. The task is then to  discriminate between threads from which the subreddit was originated. \textbf{IMDB-BINARY} is a dataset of ego-networks of actors that have appeared together in any movie.  Graphs are constructed from \textit{Action} and \textit{Romance} genres. The task is identify which genre an ego-network graph belongs to. \textbf{IMDB-MULTI} is the same, but consider three movie genres: \textit{Comedy}, \textit{Romance} and \textit{Sci-Fi}.

\subsection{Experimental setup}
In order to be as fair as possible, we follow the experimental setup of \cite{Yanardag:2015:DGK:2783258.2783417} and \cite{Shervashidze:2011:WGK:1953048.2078187} and \cite{Barnett2016}. We assess the performance of our method against some representative graph kernels, feature-based and neural networks  methods of the literature. The algorithms to which we compare are: the Graphlet \cite{Shervashidze:2011:WGK:1953048.2078187}, Shorthest path \cite{Borgwardt} and the Weisfeiler-Lehman subtree kernels \cite{Shervashidze:2011:WGK:1953048.2078187}, as well as their respective deep versions \cite{Yanardag:2015:DGK:2783258.2783417}. Random walks based kernels  as  $p-$step random walk \cite{Smola2003}, the random walk \cite{Gartner03ongraph} and Ramon \& Gartner kernels \cite{Ramon03expressivityversus} are also considered. We also compare against the feature-based method \cite{Barnett2016}, the convolutional neural network PSCN \cite{Niepert2016} and the shift aggregate extract network (SAEN) \cite{DBLP:journals/corr/OrsiniBF17}.

For the graph-kernel methods we used the Matlab scripts\footnote{http://mlcb.is.tuebingen.mpg.de/Mitarbeiter/Nino/Graphkernels/}. 
The Deep Graph kernel scripts (in Python) were taken from one of the author's website\footnote{http://www.mit.edu/$\sim$pinary/kdd/DEEP\_GRAPH\_KERNELS\_CODE.tar.gz}. We coded the feature-based approach \cite{Barnett2016} and our algorithm in Matlab. We made available our source code in this site\footnote{https://github.com/leoguti85/MaF} 
. \\

Each dataset is randomly split in training and testing sets. The best model is selected using 10-fold cross-validation with C-SVM and Random Forest. Parameter C and number of trees are optimized only on the training set. Thus, we report the generalization accuracy on the unseen test set. In order to exclude the random effect of the data splitting, we repeated the whole experiment 10 times. Finally, we report the \textit{average prediction accuracies} and its \textit{standard deviation}. \\

The parameters for the graph-kernels approaches are also cross-validated  on the training set following \cite{Shervashidze:2011:WGK:1953048.2078187} and \cite{Yanardag:2015:DGK:2783258.2783417} settings:
\begin{itemize}
\item[$\bullet$] The $p$ value for $p$-step random walk kernel is chosen from $\{1,2,3 \}$
\item[$\bullet$] We computed the random walk kernel for the decay $\lambda \in \{ 10^{-6}, 10^{-5},\ldots, 10^{-1} \}$
\item[$\bullet$] The height parameter in Ramon \& Gartner's kernel is taken from $\{1,2,3 \}$
\item[$\bullet$] In the Deep Graph kernels (GK, SP and WL), the window size and feature dimension is chosen from $\{2,5,10,25,50 \}$
\item[$\bullet$] Similar to other works \cite{Yanardag:2015:DGK:2783258.2783417}, \cite{Niepert2016} we set $h=2$ for Weisfeiler-Lehman subtree kernel
\end{itemize}

For each kernel we report the result for the parameter that achieves the best classification accuracy. For the feature-based approach \cite{Barnett2016}, feature vectors were built with the same network features they reported in their paper: number of nodes, number of edges, average degree, degree assortativity, number of triangles and global clustering coefficient. Finally, for PSCN and SAEN we compare with the accuracies reported in \cite{Niepert2016} and  \cite{DBLP:journals/corr/OrsiniBF17} respectively.

Regarding our method, Table \ref{input_features} shows the features from section \ref{features_network} we used in our experiments. The last column shows the length of the feature vector for the graphs within each dataset.

\begin{table}[!h] 
\scriptsize
\caption{Input features for each dataset \vspace*{0.5cm}}\label{input_features}
\centering
\begin{tabular}{l@{\quad}c@{\quad}c@{\quad}c@{\quad}c}
\hline
Dataset & Features & Num of eigenvectors & Node labels & Dimension  \\ 
\hline\rule{0pt}{10pt} 
MUTAG & [1,2,3,4,5,6,7] & 3 & 7 & 56  \\ [2pt]
 
PTC & [1,2,3,4,5,6,7] & 3 & 19 & 116 \\ [2pt]
 
NCI1 & [1,2,3,4,5,6,7] & 3 & 22 & 131  \\ [2pt]
 
NCI109 & [1,2,3,4,5,6,7] & 3 & 19 & 116  \\ [2pt]

ENZYMES & [1,2,3,4,5,6,7] & 3 & 3 & 36 \\ [2pt]

PROTEINS & [1,2,3,4,5,6,7] & 3 & 3 & 36 \\ [2pt]
\hline\rule{0pt}{10pt} 
COLLAB & [1,2,3,6,7] & 5 & - & 31  \\ [2pt]

REDDIT-BINARY & [1,2,3,6,7] & 5 & - & 31  \\[2pt] 

REDDIT-MULTI-5K & [1,2,3,6,7] & 5 & - & 31  \\ [2pt]

REDDIT-MULTI-12K & [1,2,3,6,7] & 5 & - & 31 \\ [2pt]

IMDB-BINARY & [1,2,3,6,7] & 3 & - & 21 \\ [2pt]

IMDB-MULTI & [1,2,3,6,7] & 3 & - & 21 \\ [2pt]
\hline \rule{0pt}{0pt} 
\end{tabular} 
\end{table}

\subsection{Statistical comparison of algorithms}\label{statistics}
We used the Friedman test \cite{Friedman1940} to compare the accuracies of different algorithms. The Friedman test is a non-parametric test based on the average ranked performances ($R_j$) of the classification performance on each dataset and is calculated as:
\begin{equation}\label{friedman}
Q = \dfrac{12D}{K(K+1)} \sum_{j=1}^K \left( R_j - \dfrac{K+1}{2} \right)^2
\end{equation}
where $D$ denotes the number of datasets, $K$ the number of algorithms and $R_j = \frac{1}{D} \sum_{i=1}^D r_i^j$ as the average rank (AR) of algorithms with $r_i^j$ denoting rank of the $j$-th of $K$ algorithms on the $i$-th of $N$ datasets. If all the algorithms perform equally well (null-hypothesis) then we can expect that $Q$ is approximately distributed as a Chi-square distribution with $K-1$ degrees of freedom. Therefore we can reject the null hypothesis and conclude that some algorithms perform better than other when $Q$ is large, with the probability that $\chi^2_{K-1} \geq Q $ as $p$-value.  

If the null-hypothesis is rejected, we can proceed with a post-hoc test. The post hoc Nemenyi test \cite{Nemenyi1963} is applied to report any significant difference between individual algorithms. The Nemenyi test states that the performance of various algorithms are significantly different if their average rank differ by at least the critical difference:

\begin{equation}\label{CD}
CD = q_{\alpha, K}\sqrt{\dfrac{K(K+1)}{12D}}
\end{equation}
where the critical values $q_{\alpha,K}$ are based on the Studentized range statistic divided by $\sqrt{2}$. Finally the results from Friedman-Nemenyi tests are displayed using the diagrams proposed by Demsar \cite{Demsar2006}. These diagrams show the ranked performances of the classification techniques along with the critical difference to stand out the algorithms which are significantly different to the best performing ones.

\subsection{Results}
We test our method on all considered benchmark datasets and compare our classification accuracies with the ones achieved by the aforementioned algorithms. We report three instances of our method: \textbf{MaF-SVM} when we train using Support Vector Machines and \textbf{MaF-RF} when we use Random Forest classifier on our multi-hop assortativities features (MaF). \textbf{MaF-nolab} corresponds to a Random Forest on a subset of features discarding explicitly node metadata information.

For the experiments on \textbf{social network} datasets we apply those algorithms that can handle unlabeled node graphs. Results for social graphs are depicted below in Table \ref{social1} (\textit{graph-kernel methods}) and Table \ref{social2} (\textit{neural nets and feature-based approach}).

\begin{table}[!h] 
\caption{\textit{Graph-kernel methods on social networks: } Mean and standard deviation of classification accuracy for Random Walk (RW) \cite{Gartner03ongraph}, Weisfeiler-Lehman (WL) \cite{Shervashidze:2011:WGK:1953048.2078187}, Graphlet (GK)\cite{5664}, Deep Graphlet kernels (DGK) \cite{Yanardag:2015:DGK:2783258.2783417},  multi-hop assortativities (MaF) (our method) }
\begin{center}\label{social1}
\scriptsize
\begin{tabular}{l@{\quad}c@{\quad}c@{\quad}c@{\quad}c@{\quad}c@{\quad}c@{\quad}}
\hline  \rule{0pt}{10pt} 
Dataset	&	RW	&	WL	&	GK	&	DGK	& MaF-SVM	& MaF-RF	\\[2pt]
\hline  \rule{0pt}{10pt} 
COLLAB	&	69.01 $ \pm $ 0.09	&	77.79 $ \pm $0.19	&	72.84 $ \pm $ 0.28	&	73.09 $ \pm $ 0.25	& 75.54 $\pm$ 1.38 &	\textbf{78.24 $ \pm $ 1.57}	\\[2pt]
IMDB-BINARY	&	64.54 $ \pm $ 1.22	&	\textbf{72.86 $ \pm $0.76}	&	65.87 $ \pm $ 0.98	&	66.96 $ \pm $ 0.56	& \textbf{71.30 $\pm$ 3.23} &  \textbf{71.60 $ \pm $ 4.45}	\\[2pt]
IMDB-MULTI	&	34.54 $ \pm $ 0.76	&	\textbf{50.55 $ \pm $0.55}	&	43.89 $ \pm $ 0.38	&	44.55 $ \pm $ 0.52	& \textbf{47.53 $\pm$ 3.24} &	45.20 $ \pm $ 3.54 \\[2pt]
REDDIT-BINARY	&	67.63 $ \pm $ 1.01	&	69.57 $ \pm $0.88	&	77.34 $ \pm $ 0.18	&	78.04 $ \pm $ 0.39	& \textbf{89.00 $\pm$ 2.25} &	\textbf{88.90 $ \pm $ 2.20}	\\[2pt]
REDDIT-MULTI-5K	&	$>$ 72h	&	47.72 $ \pm $0.48	&	41.01 $ \pm $ 0.17	&	41.27 $ \pm $ 0.18	& \textbf{54.37 $\pm$ 2.08} &	\textbf{51.39 $ \pm $ 1.91}	\\[2pt]
REDDIT-MULTI-12K	&	$>$ 72h	&	38.47 $ \pm $0.12	&	31.82 $ \pm $ 0.08	&	32.22 $ \pm $ 0.10	& \textbf{44.52 $\pm$ 1.44} &	\textbf{43.50 $ \pm $ 1.03}	\\[2pt]
\hline \rule{0pt}{10pt} 
\end{tabular}
\end{center}
\end{table}
\begin{table}
\caption{\textit{Neural nets and feature-based approaches on social networks:} Mean and standard deviation of classification accuracy for Shift agreggate extract network (SAEN) \cite{DBLP:journals/corr/OrsiniBF17}, Convolutional Neural Network (PSCN) \cite{Niepert2016}, Feature-Based (FB) \cite{Barnett2016}, multi-hop assortativities (MaF) (our method) }
\begin{center}\label{social2}
\scriptsize
\begin{tabular}{l@{\quad}c@{\quad}c@{\quad}c@{\quad}c@{\quad}c@{\quad}}
\hline  \rule{0pt}{10pt} 
Dataset	&	SAEN	&	PSCN	&	FB	&	MaF-SVM	& MaF-RF	\\
\hline  \rule{0pt}{10pt} 
COLLAB	&	75.63 $ \pm $0.31	&	72.60 $ \pm $ 2.15	&	76.35 $ \pm $ 1.64	& 75.54 $\pm$ 1.38 &	\textbf{78.24 $ \pm $ 1.57}	\\[2pt]
IMDB-BINARY	&	71.26 $ \pm $ 0.74	&	71.00 $ \pm $ 2.29	&	\textbf{72.02 $ \pm $ 4.71}	& \textbf{71.30 $\pm$ 3.23} &  \textbf{71.60 $ \pm $ 4.45}	\\[2pt]
IMDB-MULTI	&	\textbf{49.11 $ \pm $ 0.64}	&	45.23 $ \pm $ 2.84	&	47.34 $ \pm $3.56	& \textbf{47.53 $\pm$ 3.24} & 45.20 $ \pm $ 3.54 	\\[2pt]
REDDIT-BINARY	& 86.08 $ \pm $0.53	&	86.30 $ \pm $ 1.58	&	88.98 $ \pm $ 2.26	& \textbf{89.00 $\pm$ 2.25} &	\textbf{88.90 $ \pm $ 2.20}	\\[2pt]
REDDIT-MULTI-5K	& 52.24 $ \pm $0.38	&	49.10 $ \pm $ 0.70	&	50.83 $ \pm $ 1.83	& \textbf{54.37 $\pm$ 2.08} &	\textbf{51.39 $ \pm $ 1.91}	\\[2pt]
REDDIT-MULTI-12K	& \textbf{46.72 $ \pm $ 0.23}	&	41.32 $ \pm $ 0.42	&	42.37 $ \pm $ 1.27	& 44.52 $\pm$ 1.44 &	43.50 $ \pm $ 1.03 \\[2pt]
\hline 
\end{tabular}
\end{center}
\end{table}

As can be seen from Table \ref{social1} we outperform all graph kernel methods on all social networks except WL on IMDB-MULTI, while being comparable with IMDB-BINARY. In particular, our method performs better on datasets with large networks and large number of examples, see REDDIT and COLLAB in Table \ref{social1}. Regarding Table \ref{social2}, our method perform generally better than Convolutional Neural Networks (PSCN), while remains comparable with Feature Based (FB), which is expected because both are methods of the same nature.
However SAEN outperforms our method in IMDB and REDDIT multiclass problems. Experiments suggest that in general MaF-SVM is more accurate for multi-class problem than MaF-RF.

On the other hand, results of \textbf{chemoinformatic datasets} are depicted below in Table \ref{bio1-1} (\textit{Methods that do not exploit node metadata}), Table \ref{bio1-2} (\textit{random-walks and Ramon $\&$ Gartner kernel}), and Table \ref{bio1-3} (\textit{others graph-kernels and neural nets approaches})

\begin{table}[!h]
\caption{\textit{Methods that do not exploit node labels on chemoinformatic graphs: }Mean and standard deviation classification accuracy for: Graphlet (GK) \cite{5664}, Deep Graphlet (DGK)\cite{Yanardag:2015:DGK:2783258.2783417}, Feature-Based (FB) \cite{Barnett2016}) and our multi-hop assortativities Features,  without using node labels (MaF-nolab) and with labels (MaF-SVM, MaF-RF) \vspace*{0.4cm}} \label{bio1-1}
\centering
\scriptsize
\begin{tabular}{l@{\quad}c@{\quad}c@{\quad}c@{\quad}c@{\quad}c@{\quad}c@{\quad}}
\hline \rule{0pt}{10pt} 
Data	&	GK	&	DGK	&	FB	&	MaF-nolab	&	MaF-SVM	& MaF-RF\\[2pt]
\hline \rule{0pt}{10pt} 
MUTAG	&	81.66 $ \pm $ 2.11	&	82.66 $ \pm $ 1.45	&	84.66 $ \pm $ 2.01	&	82.48 $ \pm $  9.28	& \textbf{85.09 $ \pm $ 7.34} &	\textbf{89.89 $ \pm $ 5.58}	\\[2pt]
PTC	&	57.26 $ \pm $ 1.41	&	57.32 $ \pm $1.13	&	55.58 $ \pm $ 2.30	&	\textbf{61.99 $ \pm $  7.06}	& \textbf{58.79 $ \pm $ 7.11} &	\textbf{61.34 $ \pm $ 7.61}	\\[2pt]
NCI1	&	62.28 $ \pm $ 0.29	&	62.48 $ \pm $ 0.25	&	62.90 $ \pm $ 0.96	&	\textbf{70.12 $ \pm $ 1.58}	& \textbf{73.89 $ \pm $ 1.48} &	\textbf{77.32 $ \pm $ 1.68}	\\[2pt]
NCI109	&	62.60 $ \pm $ 0.19	&	62.69 $ \pm $ 0.23	&	62.43 $ \pm $ 1.13	&	\textbf{67.87 $ \pm $  2.15}	& \textbf{73.49 $ \pm $ 1.82} &	\textbf{74.97 $ \pm $ 2.19}	\\[2pt]
PROTEINS	&	71.67 $ \pm $ 0.55	&	71.68$ \pm $ 0.50	&	69.97 $ \pm $ 1.34	&	\textbf{73.05 $ \pm $  3.35}	& \textbf{75.20 $ \pm $ 2.67} &	\textbf{76.73 $ \pm $ 2.97}	\\[2pt]
ENZYMES	&	26.61 $ \pm $ 0.99	&	27.08 $ \pm $ 0.79	&	29.00 $ \pm $ 1.16	&	\textbf{40.67 $ \pm $  3.96}	& \textbf{48.17 $ \pm $ 6.60} &	\textbf{56.17 $ \pm $ 9.10}	\\[2pt]
\hline
\end{tabular}
\end{table}
\begin{table}[!h]
\caption{\textit{Random-walks and Ramon $\&$ Gartner kernels on chemoinformatic graphs: }Mean and standard deviation classification accuracy for: Ramon \& Gartner (RG) \cite{Ramon03expressivityversus}, p-Random Walk (pRW)  \cite{Smola2003}, Random Walk (RW) \cite{Gartner03ongraph}) and our multi-hop assortativities,  without using node labels (MaF-nolab) and with labels (MaF-SVM, MaF-RF) \vspace*{0.4cm}} \label{bio1-2}
\centering 
\scriptsize
\begin{tabular}{l@{\quad}c@{\quad}c@{\quad}c@{\quad}c@{\quad}c@{\quad}c@{\quad}} 
\hline \rule{0pt}{10pt} 
Data	&	RG	&	pRW	&	RW	&	MaF-nolab	&	MaF-SVM	& MaF-RF\\[2pt]
\hline \rule{0pt}{10pt} 
MUTAG	&	84.88 $ \pm $  1.86	&	80.05 $ \pm $  1.64	&	83.72 $ \pm $  1.50	&	82.48 $ \pm $  9.28	& \textbf{85.09 $ \pm $ 7.34} &	\textbf{89.89 $ \pm $ 5.58}	\\[2pt]
PTC	&	58.47 $ \pm $  0.90	& 59.38 $ \pm $  1.66	&	57.85 $ \pm $  1.30	&	\textbf{61.99 $ \pm $  7.06}	& 58.79 $ \pm $ 7.11 &	\textbf{61.34 $ \pm $ 7.61}	\\[2pt]
NCI1	&	56.61 $ \pm $  0.53	&	$>$ 72h	&	48.15 $ \pm $  0.50	&	\textbf{70.12 $ \pm $ 1.58}	& \textbf{73.89 $ \pm $ 1.48} &	\textbf{77.32 $ \pm $ 1.68}	\\[2pt]
NCI109	&	54.62 $ \pm $  0.23	&	$>$ 72h	&	49.75 $ \pm $  0.60	&	\textbf{67.87 $ \pm $  2.15}	& \textbf{73.49 $ \pm $ 1.82} &	\textbf{74.97 $ \pm $ 2.19}	\\[2pt]
PROTEINS	&	70.73 $ \pm $  0.35	&	71.16 $ \pm $  0.35	&	74.22 $ \pm $  0.42	&	73.05 $ \pm $  3.35	& \textbf{75.20 $ \pm $ 2.67} &	\textbf{76.73 $ \pm $ 2.97}	\\[2pt]
ENZYMES	&	16.96 $ \pm $  1.46	&	30.01 $ \pm $  1.01	&	24.16 $ \pm $  1.64	&	\textbf{40.67 $ \pm $  3.96}	& \textbf{48.17 $ \pm $ 6.60} &	\textbf{56.17 $ \pm $ 9.10}	\\[2pt]
\hline
\end{tabular}
\end{table}
\begin{table}[!h] 
\caption{\textit{Other graph-kernels and neural nets approaches on chemoinformatic graphs: }Deep Shortest-path (DSP) \cite{Yanardag:2015:DGK:2783258.2783417}, Weisfeiler-Lehman (WL) \cite{Shervashidze:2011:WGK:1953048.2078187}, Deep Weisfeiler-Lehman (DWL) \cite{Yanardag:2015:DGK:2783258.2783417} kernels, Convolutional Neural Network (PSCN) \cite{Niepert2016}) and multi-hop assortativities (MaF-SVM, MaF-RF) (our method) \vspace*{0.4cm}}\label{bio1-3}
\scriptsize 
\centering 
\begin{tabular}{l@{\quad}c@{\quad}c@{\quad}c@{\quad}c@{\quad}c@{\quad}c@{\quad}} 
\hline \rule{0pt}{10pt} 
Data	&	DSP	&	WL	&	DWL	&	PSCN	&	MaF-SVM	& MaF-RF\\[2pt]
\hline \rule{0pt}{10pt} 
MUTAG	&	87.44 $ \pm $ 2.72	&	80.72 $ \pm $ 3.00	&	82.94 $ \pm $ 2.68	&	\textbf{92.63 $ \pm $ 4.21}	& 85.09 $ \pm $ 7.34 &	\textbf{89.89 $ \pm $ 5.58}	\\[2pt]
PTC	&	59.52 $ \pm $ 2.19	&	56.97 $ \pm $ 2.01	&	59.17 $ \pm $ 1.56	&	60.00 $ \pm $ 4.82	& 58.79 $ \pm $ 7.11 &	\textbf{61.34 $ \pm $ 7.61}	\\[2pt]
NCI1	&	73.55 $ \pm $ 0.51	&	80.13 $ \pm $ 0.50	&	\textbf{80.31 $ \pm $ 0.46}	&	78.59 $ \pm $ 1.89	& 73.89 $ \pm $ 1.48 &	77.32 $ \pm $ 1.68	\\[2pt]
NCI109	&	73.26 $ \pm $ 0.26	&	80.22 $ \pm $ 0.34	&	\textbf{80.32 $ \pm $ 0.33}	&	--	& 73.49 $ \pm $ 1.82 &	74.97 $ \pm $ 2.19	\\[2pt]
PROTEINS	&	75.78 $ \pm $ 0.54	&	72.92 $ \pm $ 0.56	&	73.30 $ \pm $ 0.82	&   75.89 $ \pm $ 2.76	& 75.20 $ \pm $ 2.67 &	\textbf{76.73 $ \pm $ 2.97}	\\[2pt]
ENZYMES	&	41.65 $ \pm $ 1.57	&	53.15 $ \pm $ 1.14	&	53.43 $ \pm $ 0.91	&	--	& 48.17 $ \pm $ 6.60 &	\textbf{56.17 $ \pm $ 9.10}	\\[2pt]
\hline
\end{tabular}
\end{table}
In Table \ref{bio1-1} we compare against methods cannot handle node metadata. As we can see, our method (MaF-nolab) based solely on the structure of the graphs outperforms in all datasets to GK, DGK and FB algorithms. Similarly, comparing against graph kernel methods in Table \ref{bio1-2}, our approach exhibits the best performance along all considered datasets.  Even in the case where we do not take node metadata into account our approach has remarkable performances. Indeed, including node metadata into our algorithm improves in general our classification accuracy. We see also in Table \ref{bio1-3} that in the multi-class dataset (ENZYMES) our approach performs the best being consistent with the results on social networks. Looking in particular at the proteins graphs (PTC, PROTEINS), our method outperforms other approaches. Meanwhile, unlike for social network datasets our method is outperformed by the deep WL and WL kernels on NCI graphs see Table \ref{bio1-3}.

We used the statistical significance analysis introduced on section \ref{statistics} to compare the accuracies of the different algorithms. We applied it on chemoinformatic and social networks datasets independently, including only algorithms with complete scores along datasets, e.g full-filled columns on Tables \ref{social1}, \ref{social2}, \ref{bio1-1}, \ref{bio1-2} and \ref{bio1-3}.

For each group, the Friedman chi-square statistic (Eq. \ref{friedman}) and corresponding $p$-values were computed. Indeed, $Q = 34.63$ with corresponding $p$-value of $0.000144$ are reported for chemoinformatic datasets and $Q = 24.66$ with $p$-value of $0.000869$ for the social network benchmark.  

\begin{figure}[!ht]
\centering{\includegraphics[width=5.4in]{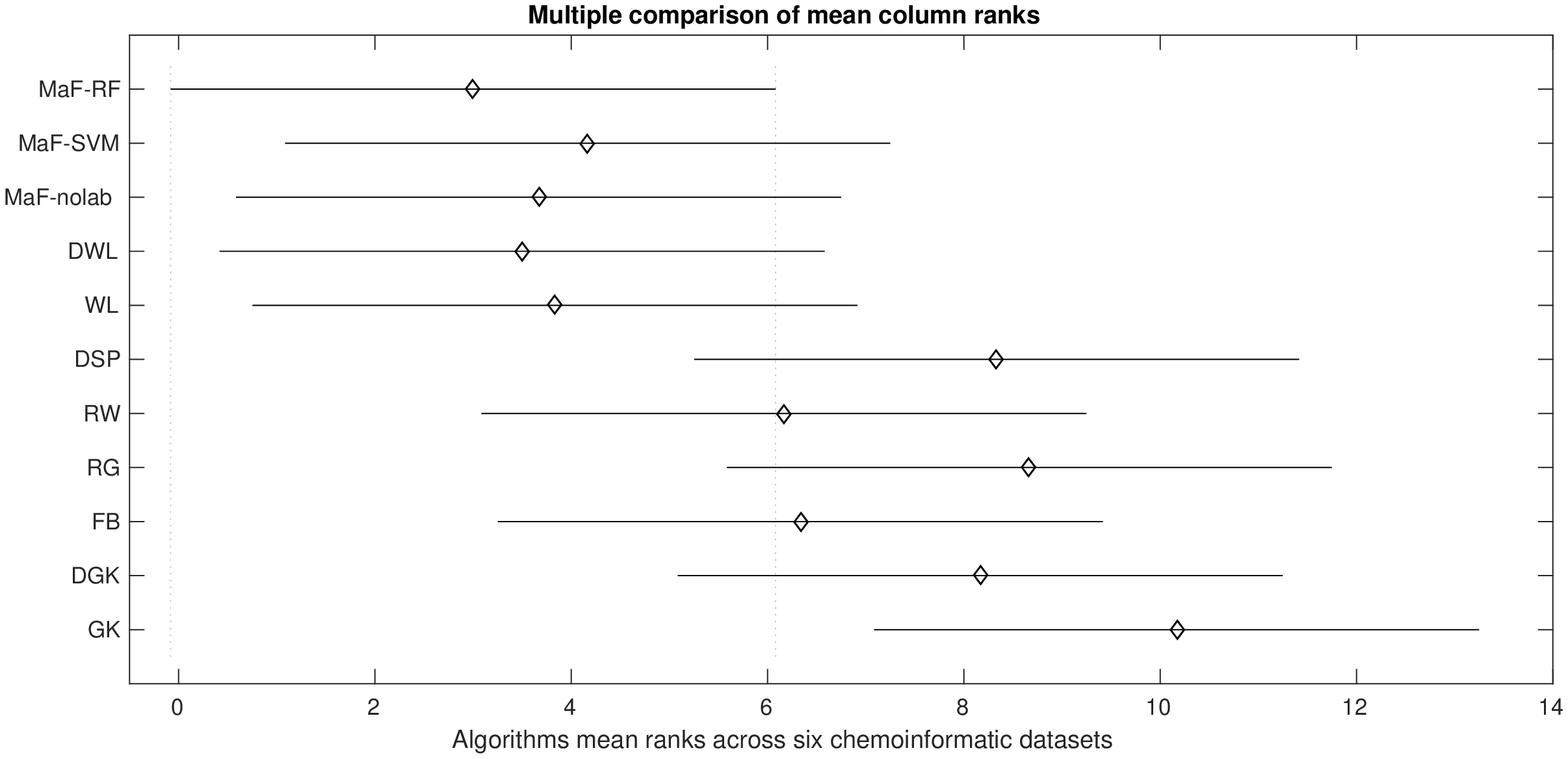}}
\caption{Significance diagram on chemoinformatic datasets} 
\label{chemo_ranks}
\end{figure}

\begin{figure}[!ht]
\centering{\includegraphics[width=5.4in]{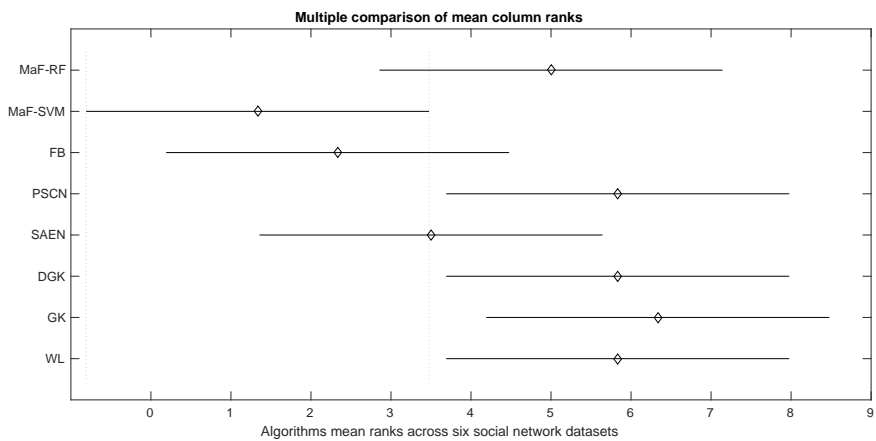}}
\caption{Significance diagram on social network datasets} 
\label{social_ranks}
\end{figure}

As these indicate that some algorithms have difference performances than others ($p < 0.005$) a post hoc Nemenyi test was applied on each class distribution. The following significance diagrams (Figures \ref{chemo_ranks} and \ref{social_ranks}) display the accuracy performance ranks of the algorithms along with the Nemenyi's critical difference (CD) tail (Eq \ref{CD}). The CD value for social network benchmarks was $4.2863$ and $6.1633$ for chemoinformatics. Each diagram shows the algorithms used on each benchmark, listed in ascending order of ranked performance on the $y-$axis, and the algorithm's mean rank across all six datasets (Table \ref{datasets}), displayed on $x-$axis. Two vertical dashed lines were inserted indicating the start and the end of the best performing method.

Regarding chemoinformatic benchmarks in the Figure \ref{chemo_ranks}, it shows that our method MaF-RF is the best performing technique with an average ranking of $3.0$. The diagram clearly shows that Graphlet kernel (GK) perform significantly worst than the best performing algorithm, with an AR of $10.166$. We can confirm our previous observation in which exploiting node attributes improves the performance of the method.

In relation to the social network benchmarks, Figure \ref{social_ranks} shows that our MaF-SVM method perform the best with an AR of $1.333$. This diagram clearly shows that Weisfeiler-Lehman (WL), Graphlet (GK), Deep Graphlet kernels as well as the convolutional neural network (PSCN) perform significantly worse than the best performing technique, with AR values of $5.833$, $6.333$, $5.833$ and $5.833$ respectively.


\subsection{Computational Cost}
Computing our assortativity features relies principally in two aspects: eigenvector computations and matrix-vector multiplication (eg. covariance matrix times eigenvectors). In order to make fair comparisons, for all methods  we report the runtime for features generation. Therefore, we report the runtime for graph-kernel matrix computation and the multi-hop assortativities feature vector generated with the attributes enumerated in section \ref{features_network}. The results are shown in Table \ref{exetime}.

\begin{table}[!h]
\scriptsize
\caption{Runtime for features and kernels computations}\label{exetime}
\begin{center}
\begin{tabular}{l@{\quad}c@{\quad}c@{\quad}c@{\quad}c@{\quad}c@{\quad}c@{\quad}c@{\quad}}
\hline  \rule{0pt}{10pt} 
Dataset & MaF & SP & DSP & GK & DGK & WL & DWL  \\ 
\hline \rule{0pt}{10pt} 
MUTAG & 0.64 s  & 0.67 s &  0.68 s & 8.7 s & 8.03 s & 0.48 s & 0.19 s  \\ [2pt]
PTC   & 1.64 s  & 3.48 s &  3.48 s & 14.04 s & 13.5 s &  1.12 s & 0.93 s  \\ [2pt]
NCI1  & 26.34 s & 45.78 s & 50.11 s & 2.8 min & 2.7 min &  17.36 & 13.20 s \\ [2pt]
NCI109 & 27.14 s & 43.44 s &  44.91 s & 2.7 min & 2.7 min &  17.58 s & 13.65 s \\ [2pt]
PROTEINS & 9.12 s & 39.75 &  40.29 s & 46.8 s & 42.7 s &  5.21 s & 16.9 s \\ [2pt]
ENZYMES  & 3.91 s & 6.53 & 6.27 s & 26.05 s & 26.17 s &  2.61 s & 4.08 s \\ [2pt]
\hline \rule{0pt}{10pt} 
COLLAB & 7.14 min & NA &  NA & 5.01 min & 4.91 min &  54.4 s &  ME \\ [2pt]
IMDB-BINARY & 8.88 s & NA &  NA & 46.4 s & 44.60 s &  2.46 s & 2.05 s  \\[2pt] 
IMDB-MULTI & 5.59 s & NA &  NA & 48.12 s & 46.7 s &  2.37 s & 1.91 s  \\ [2pt]
REDDIT-BINARY & 14.40 min & NA &  NA & 1.4 min & 1.4 min &  1.71 min & ME \\ [2pt]
REDDIT-MULTI-5K & 29.55 min & NA &  NA & 3.8 min & 3.7 min &  5.98 min & ME  \\ [2pt]
REDDIT-MULTI-12K & 58.99 min & NA &  NA & 8.7 min & 8.4 min &  14.4 min & ME  \\ [2pt]
\hline \rule{0pt}{0pt} 
\end{tabular} 
\end{center}
\end{table}
 
Our method (MaF) is clearly faster than shortest path (SP) and deep shortest path (DSP) kernels which do not work on unlabeled node networks (NA). We also outperform Graphlet (GK) and Deep Graphlet (DGK) kernels on chemoinformatic and small size datasets. Among the graph-kernel methods Weisfeiler-Lehman (WL) is the fastest one, scaling very well on big datasets but at the expense of losing accuracy capabilities (Table \ref{social1}, \ref{bio1-3}). Deep WL kernel remains a fast algorithm in small size graphs, e.g chemoinformatic datasets, but requires huge memory (ME) for larger social networks, on which our method works well. Although our method shows longer run times on REDDIT datasets due to the eigenvectors computation performed on large networks, we still keep much better classification accuracies than other methods (Tables \ref{social1} and \ref{social2}).

It is worth mentioning that the computation of multi-hop assortativities can be performed by the means of Monte Carlo estimates on simulated random walks instead of the exact linear algebra formula (Eq. \ref{vector_covariance}). This can be advantageous when the number of nodes and edges in the network, and the number of hops, preclude efficient matrix-vector computations. Note that in these circumstances, computing second, third or other eigenvectors will also likely prove unconvenient. None of the benchmarks used in this article present these characteristics. 

All our computations were done on a standard computer Intel(R) Core(TM) i7-4790 CPU, 3.60GHzI and 16G of RAM. The algorithms that exhausted our memory capabilities were flagged as ME (memory error) in the previous table.

\section{Discussion}
In this work we introduce an extension of the intuitive notion of assortativity for networks and a particular application in graph classification. We define multi-hop assortativities by setting up a dynamic on the network and computing covariances over diverse node attributes among multiples time scales. It turns out that those features on a reduced number of structural attributes in addition to node metadata whenever present, are useful to characterize networks. We use these assortativities in the context of networks classification. The classification is performed by training a Support Vector Machine and Random Forest classifiers, achieving high accuracies on both social and biochemical datasets. The Friedman and Nemenyi post hoc tests were then used to determine whether the differences between the average ranked accuracies were statistically significant. The experimental results reveal that our approach is particularly effective  when it is applied on large networks and datasets of many (possibly small) graphs, performing significantly better than kernel methods and convolutional neural networks. However, we show competitive accuracies when it is applied on small graphs with almost zero clustering coefficient such as graphs representing molecules or proteins. When node metadata is available, our method outperforms kernel-methods and random walks based baselines and, remains competitive against neural networks approaches.

It is worth mentioning that although we experimented with networks with a single categorical node attribute, our approach is applicable to networks with scalar and multiple attributes, e.g a social network where nodes attributes are age, weight, gender, etc.

Our framework can also be tuned to incorporate the assortivities of any node attribute we deem relevant to the application, for instance betweenness centrality, etc. Instead of the simple random walk, we may also use an application-specific diffusion dynamics, eg in continuous-time, biased towards some nodes, etc. \cite{Lambiotte2014a,MASUDA20171}. In this proof-of-concept paper, we only use the simplest dynamics, most elementary attributes with few eigenvectors, number of nodes and edges.

An example of application for future work is on brain networks (connectome datasets) \cite{hagmann2008mapping, B.Chiem2018}, where discriminating between healthy and non-healthy connectomes is a task of growing importance.



\section*{Funding}
This  work  was  supported  by  Concerted  Research
Action (ARC) supported by the Federation Wallonia-Brussels Contract ARC
14/19-060 and Flagship European Research Area  Network  (FLAG-ERA) Joint  Transnational  Call  “FuturICT  2.0” to which are gratefully acknowledged.

\section*{Acknowledgment}
We thank Marco Saerens, Leto Peel and Roberto D'Ambrosio for helpful discussions and suggestions.


%


\bibliographystyle{imaiai}
\ifx\undefined\BySame
\newcommand{\BySame}{\leavevmode\rule[.5ex]{3em}{.5pt}\ }
\fi
\ifx\undefined\textsc
\newcommand{\textsc}[1]{{\sc #1}}
\newcommand{\emph}[1]{{\em #1\/}}
\let\tmpsmall\small
\renewcommand{\small}{\tmpsmall\sc}
\fi

\end{document}